\documentclass[10pt,conference]{IEEEtran}
\usepackage{amsmath,amssymb,mathtools,bm,etoolbox}
\usepackage{blindtext}
\usepackage{graphicx}
\usepackage{array,tabularx}
\usepackage{bigstrut}
\usepackage{booktabs}
\usepackage{tikz}
\usepackage{graphicx,adjustbox}
\usepackage{pgfplots}
\usepackage{tabulary}
\newenvironment{conditions*}
  {\par\vspace{\abovedisplayskip}\noindent
   \tabularx{\columnwidth}{>{$}l<{$} @{${}={}$} >{\raggedright\arraybackslash}X}}
  {\endtabularx\par\vspace{\belowdisplayskip}}
\usepackage{subcaption}
\usepackage{caption}
\usepackage{kantlipsum} 
\usepackage{mwe} 
\usepackage[noadjust]{cite}
\DeclareCaptionLabelSeparator{periodspace}{.\quad}
\usepackage{algorithm,float}
\usepackage{algorithmic}
\newcommand\ddfrac[2]{\frac{\displaystyle #1}{\displaystyle #2}}
\usepackage{pgfplots}
\usepackage{tikz}
\usetikzlibrary{patterns}

\usepackage[justification=centering]{caption}

\usepackage{authblk}

\ifCLASSINFOpdf
\else
\fi

\usepackage[english]{babel}
\usepackage[utf8]{inputenc}
\usepackage{graphicx,color}
\usepackage{booktabs}
\usepackage{multirow} 

\usepackage{tikz}

\begin{document}
%
\title{VANETs Meet Autonomous Vehicles: \\A Multimodal 3D Environment Learning Approach}
\author{Yassine Maalej, Sameh Sorour, Ahmed Abdel-Rahim and Mohsen Guizani \\
        University of Idaho, Moscow, ID, USA         }
        
\maketitle
\begin{abstract}
In this paper, we design a multimodal  framework for object detection, recognition and mapping based on the fusion of stereo camera frames, point cloud Velodyne Lidar scans, and Vehicle-to-Vehicle (V2V) Basic Safety Messages (BSMs) exchanged using Dedicated Short Range Communication (DSRC). We merge the key features of rich texture descriptions of objects from 2D images, depth and distance between objects provided by 3D point cloud and awareness of hidden vehicles from BSMs' 3D information. We present a joint pixel to point cloud and pixel to V2V correspondences of objects in frames from the Kitti Vision Benchmark Suite by using a semi-supervised manifold alignment approach to achieve camera-Lidar and camera-V2V mapping of their recognized objects that have the same underlying manifold.  
\end{abstract}
\IEEEpeerreviewmaketitle
\section{Introduction}
For years, researchers on Vehicular Ad-hoc Networks (VANETs) and Autonomous vehicles presented various solutions for vehicle safety and automation, respectively. Yet, the  developed works in these two areas have been mostly conducted in their own separate worlds, and barely affected one-another despite the obvious relationships. The US National Science Foundation and US Department of Transportation have expressed tremendous need and importance to relate these two worlds together in several of their call for research proposal in 2017 \cite{NSFCyberPhysicalSystems}. They clearly emphasized on the major importance of integration and fusion of data from various input modes in order to create a deeper understanding of object surroundings. Precisely, enriched 3D scene reconstruction by different input technologies and deep learning techniques are of a paramount importance to allow automated vehicle system to perform effectively and safely on roads. These directions are strongly supported by the multiple accidents and traffic light violations made by autonomous vehicle prototypes from top players in the market (e.g., Tesla, Uber), incidents that could have easily been mitigated if communications among vehicles and to the road infrastructure would have been considered \cite{UberCrash, UberBlowsThroughRedLight}. 
To respond to this need, we propose in this paper to enrich the learning of the 3D vehicle surroundings using multimodal inputs, namely Lidar scans, camera frames, V2V-conveyed basic safety messages (BSMs). Learning the correspondence between the same objects from different data inputs is a difficult but necessary task that self-driving vehicles have to handle especially with the curse of the data representation and dimensionality. Incorporating objects detected from these three sources is a mapping process that can be casted as a manifold alignment problem \cite{SemisupervizedAlignmentOfmanifolds}. 
3D point cloud is omnipresent in free space, obstacle detection and avoidance, path planning and in autonomous driving systems. 2D convolutional neural networks (CNNs) \cite{fusonNet2D} for processing 3D Lidar point cloud is widely used for point cloud data recognition. However, it is not considered an optimal solution since it requires a model to recover the original geometric relationships. Vote3Deep is developed in \cite{Vote3Deep} for fast point cloud object detection from 3D CNN in order to keep the key power of Lidar as distance and objects 3D shapes and depth. The Kitti Vision Benchmark Suite \cite{VisionMeetsRoboticsTheKITTIDataset} offers raw Lidar and labeled objects from point cloud as well as RGB and Grayscale sequences images of driving. However, it does not offer any V2V type of message \cite{DesignOf59GHzDSRCBasedVehicularSafetyCommunication,AdvancedActivityAwareMultichannelOPerations16094} which offers various types of safety applications operating on a control channel of its 7 available channels operating over a dedicated 75 MHz spectrum band around 5.9 GHz. The goal of this work is to merge the key features of Lidar in giving accurate distances, camera with object textural details, and V2V beacons for the awareness of both hidden out-of-sight vehicles or vehicles not observed by the two other means due to bad conditions (e.g., rainy or foggy weather). Our Framework requires additional prior knowledge about selection of labeled paired objects between the 3 types of data set that we want to correspond. Exploring the physical neighborhood correlation within these three datasets and their natural correspondences in the 3D physical space, we cast the merging problem of these three sets of data as a semi-supervised manifold alignment. Given some clear correspondences between data points from each pair of data sets, we align (i.e., pair) the rest of the points between the camera-lidar and camera-V2V data sets. The problem is casted as an eigenvalue problem over a graph-based compounded Laplacian matrix. Once the mapping of known points is done, the other points from each data sets can be easily added in aligned 3D environment, thus significantly enriching the vehicle knowledge of its surroundings.
The remainder of this paper is organised as follows. The related work is presented in Section \ref{RelatedWorkAndMotivation}. Object recognition in scene of the kitti dataset are presented in \ref{AdaptedDarknetConvolutionalNeuralNetwork}. Learning the Lidar objects from Lidar point cloud scans in Kitti dataset are studied in \ref{3DObjectsDetectionFromPointCloudLidarData}. We present the manifold alignment formulation and solution between the 3 Dimensional Lidar space, D camera Space, and 3D V2V becons in section \ref{SemiSupervisedAlignmentOfManifolds}. BSM creation according to the Lidar recognized objects from Kitti suite, number of recognized objects per input type and the performance of the alignment process are illustrated in Section \ref{NumericalAnalysisOfMappingAccuracyAndErrors}. Section \ref{Conclusion} provides conclusion and future work. 
\section{Related Work}\label{RelatedWorkAndMotivation}
One approach to understand the scene around vehicles is semantic segmentation that labels each pixel in an image with the category of the belonging objects. Labeling each pixel of the scene independently from its surrounding pixels is a very hard task to achieve. In order to know the category of a pixel, we have to rely on relatively short-range surrounding information and long-range information. In other words, to determine that a certain pixel belongs to a vehicle, a person or to any other class of objects, we need to have a contextual window that is wide enough to show the surrounding of the pixel and consequently to make an informed decision of the object class that contains the pixel.  
Techniques based on Markov Random Fields (MRF), Conditional Random Field (CRF) and many graphical models are presented in 
\cite{LearningHybridModelsForImageAnnotationwithPartiallyLabeledData}, 
\cite{AssociativeHierarchicalCRFsForObjectClassImageSegmentation}, \cite{StackedHierarchicalLabeling} to guarantee the consistency of labeling of the pixels in the context of the overall image. In addition, the authors in \cite{SuperparsingScalableNonparametricImageParsingWithSuperpixels}, \cite{EfficientlySelectingRegionsForSceneUnderstanding} and \cite{DecomposingASceneIntoGeometricAndSemanticallyConsistentRegions} developed various methods for presegmentation into superpixels or segment candidates that are used to extract the categories and features characterizing individual segments and from combinations of neighboring segments. Alternatively,  the authors in \cite{FeatureSpaceOptimizationForSemanticVideoSegmentation} attempted to create 3D reconstruction of dynamic scenes by achieving a long-range spatio-temporal regularization in semantic video segmentation, since both the camera and the scene are in motion. The developed idea is to integrate deep CNN and CRF to perform sharp pixel-level boundaries of objects. To this end, deep learning has shown the best performance in inferring objects from not previously trained or seen scenes. Joseph \textit{et al.} \cite{YOLODARKNET} developed a general purpose object detection system characterized by a resolution classifier and the usage of a 2 fully connected networks that are built on top of a 24 convolutional layers network.  Additionally, a unified muti-scale deep CNN for real-time object detection is developed in \cite{mscnn} with many sub-network detectors with multiple output layers for multiple object class recognition. Most autonomous driving systems rely on Lidar, stereo cameras or radar sensors to achieve object detection, scene flow estimation of objects on roads and their key characteristics and influence on driving decisions and steering commands. We present an augmented scene flow understanding and object mapping by considering not only Lidar and cameras, but also DSRC-based V2V beacons exchanged between vehicles. 
\section{Adapted Darknet's Convolutional Neural Network  and KITTI frame testing} \label{AdaptedDarknetConvolutionalNeuralNetwork}
\begin{center}
    \begin{figure*} 
      \includegraphics[width=1.0\textwidth, height=6cm]{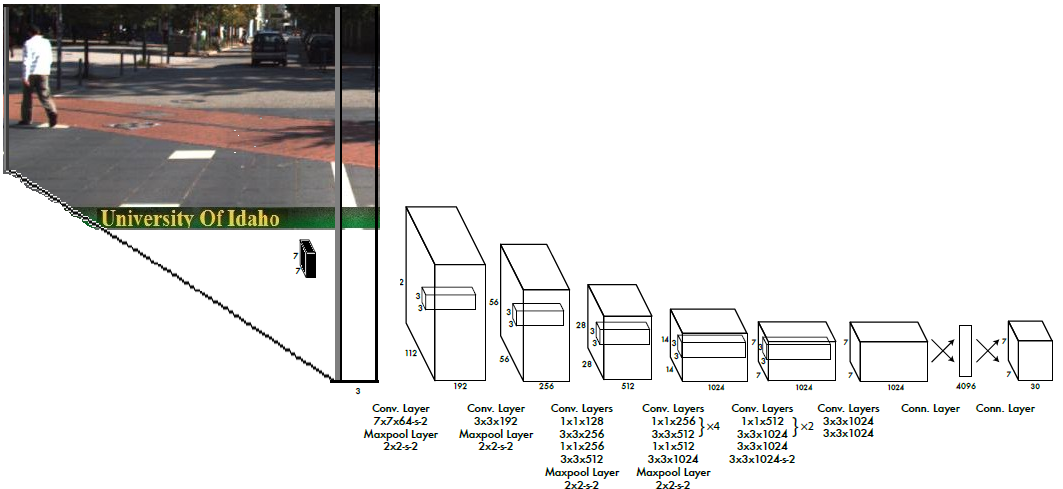}
      \caption{Darknet Convolutional Neural Network Architecture \cite{YOLODARKNET}}
      \label{DarknetYolo}
    \end{figure*}
\end{center}

\begin{figure}[H]
\begin{center} 
\includegraphics[width=9cm,height=4cm]{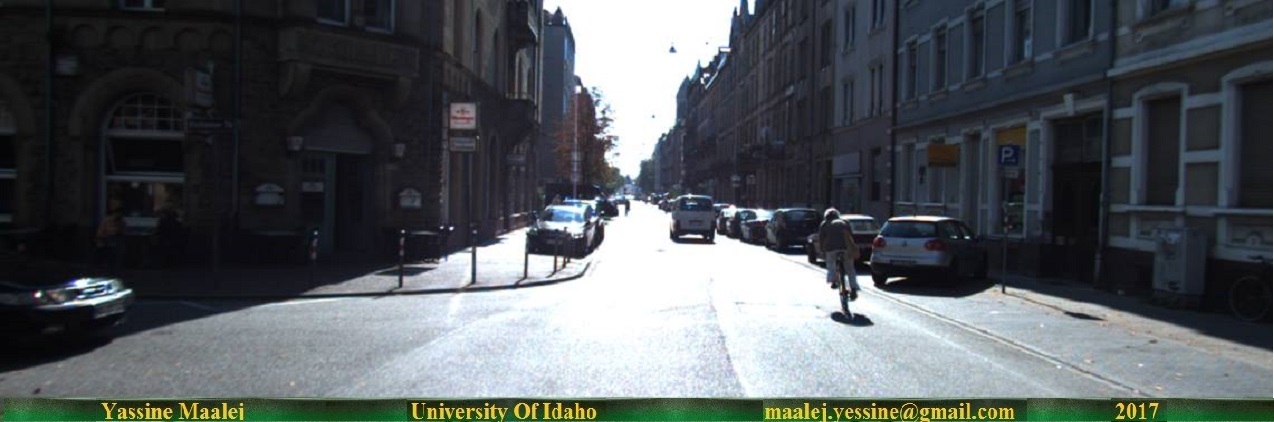}
\end{center}
\caption{Frame$(a)$ of RGB sequence drive 2011\_09\_26 drive\_0005 \#117}
\label{FrameA}
\end{figure}

\begin{figure}[H]
\begin{center} 
\includegraphics[width=9cm,height=4cm]{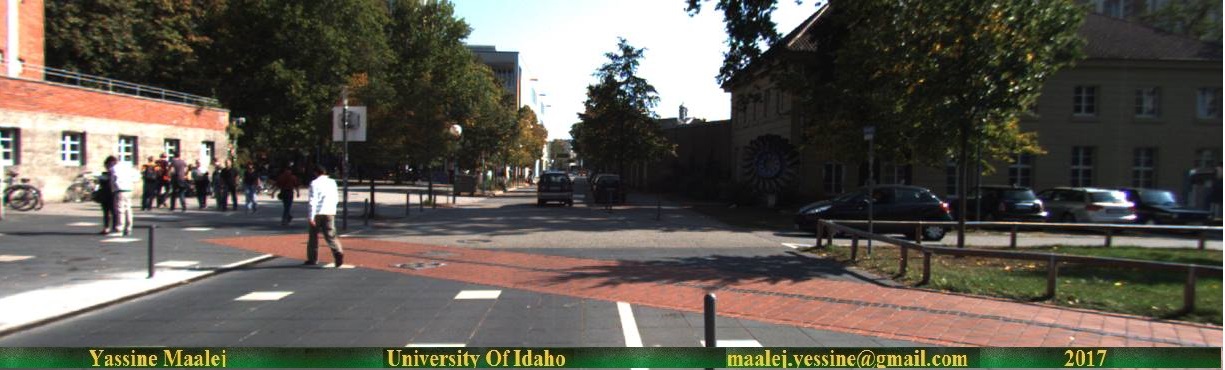}
\end{center}
\caption{Frame$(b)$ of RGB sequence drive 2011\_09\_28 drive\_00016 \#32}
\label{FrameB}
\end{figure}

Inspired by CNN developed in \cite{YOLODARKNET}, we propose to exploit the feature of Anchor Boxes that predict the coordinates of the bounding boxes around recognized objects to find their pixel adjacency directly from the fully connected layers that are developed on top of the convolutional network extractor, as described in Fig. \ref{DarknetYolo}.  Fig. \ref{FrameA} and Fig. \ref{FrameB} represent the original frames of the two different driving sequences from Kitti. Both of frame$(a)$ and frame$(b)$ present random object count per class either for image recognition or for labeled Lidar objects.  

\begin{figure}[H]
\begin{center} 
\includegraphics[width=9cm,height=4cm]{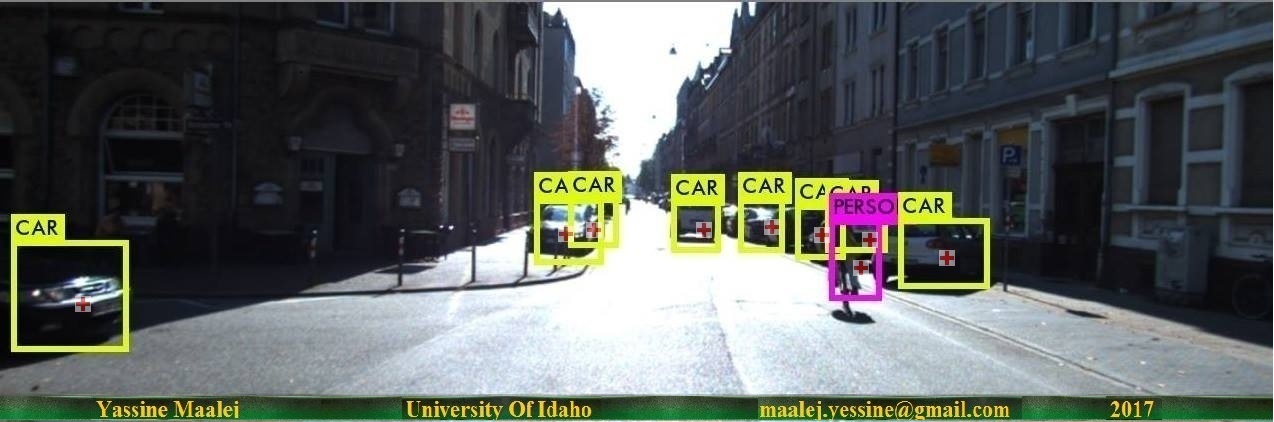}
\end{center}
\caption{Frame$(a)$ Object Detection and moment calculus}
\label{FrameA_object_detection}
\end{figure}

\begin{figure}[H]
\begin{center} 
\includegraphics[width=9cm,height=4cm]{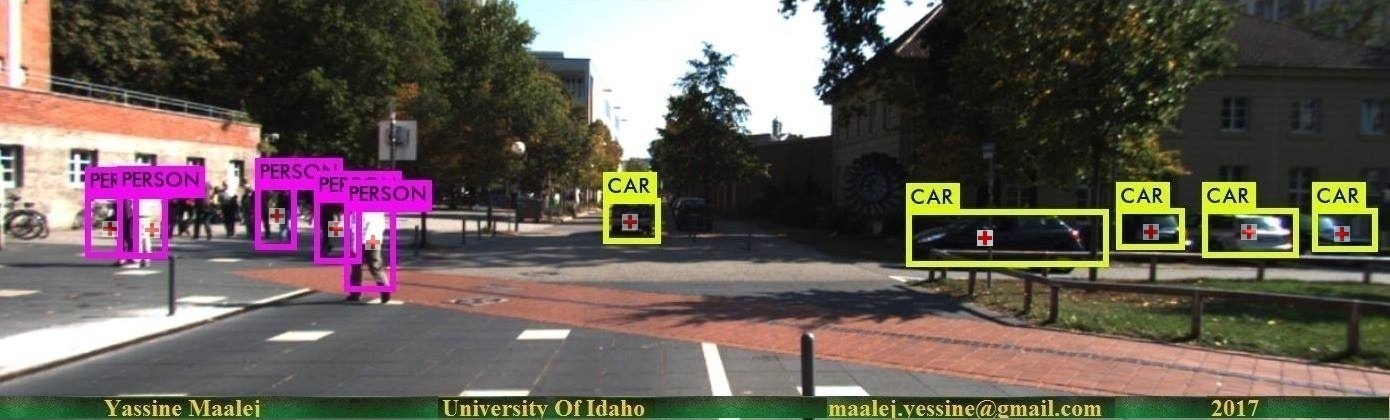}
\end{center}
\caption{Frame$(b)$ Object Detection and moment calculus}
\label{FrameB_object_detection}
\end{figure}

Fig. \ref{FrameA_object_detection} and Fig. \ref{FrameB_object_detection} represent the detected and recognized objects in frame$(a)$ and frame$(b)$ respectively. Each detected object's moment is the same as its surrounding box and is expressed in pixels.  We notice that in Fig. \ref{FrameA_object_detection}, some vehicles are not detected in the right side of the parked vehicles.  

In Fig. \ref{FrameB_object_detection}, our detection missed the vehicle next to the one in the background of the image. We will overcome the problem of undetected objects by the alignment and we will no longer have a missing knowledge of surrounding objects.

\begin{figure}[H]
\begin{center} 
\includegraphics[width=9cm,height=6cm]{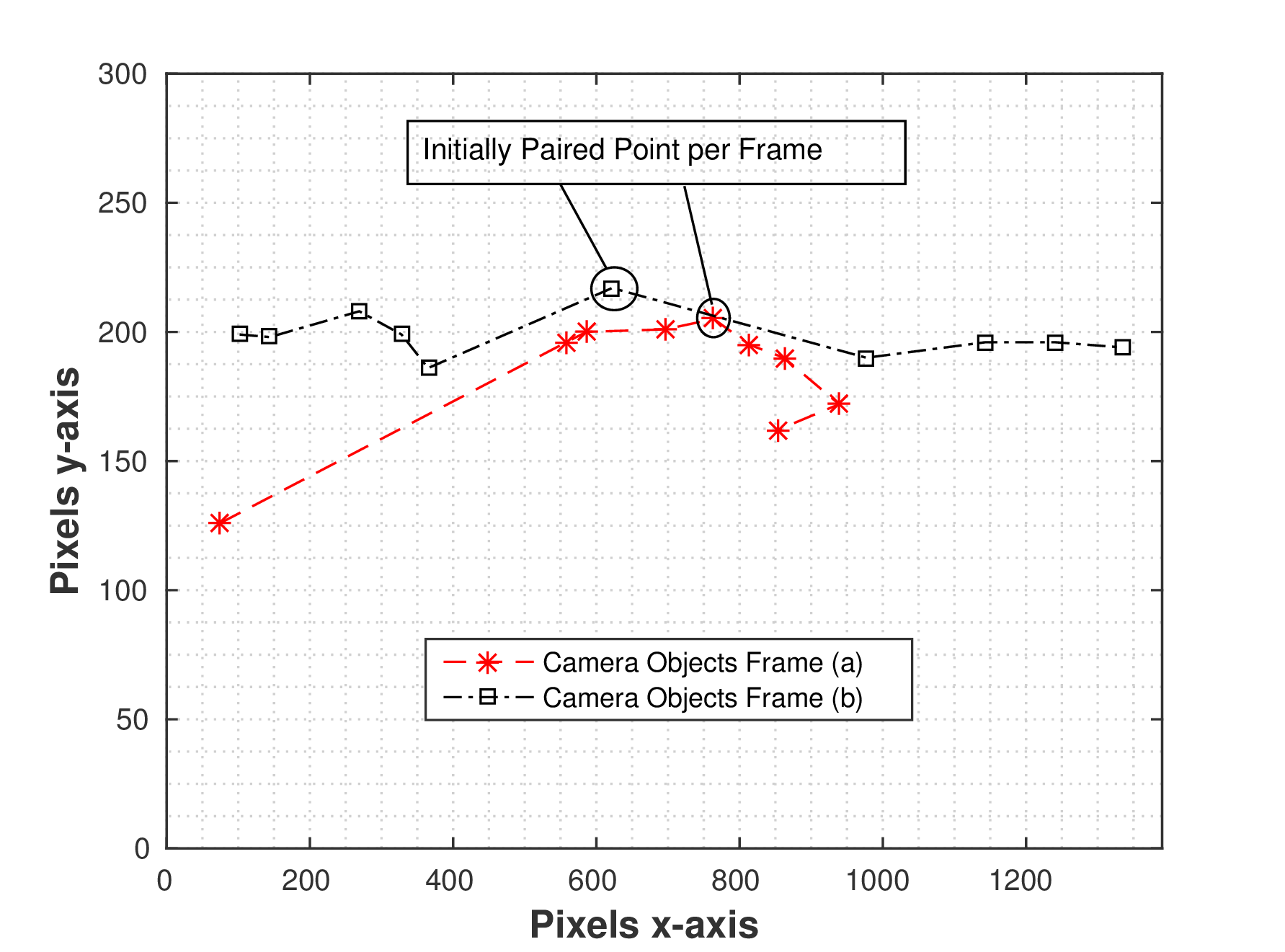}
\end{center}
\caption{Pixel-wise adjacency of recognized objects}
\label{PixelWiseAdjacencyOfRecognizedObjects}
\end{figure}

In Fig. \ref{PixelWiseAdjacencyOfRecognizedObjects}, we draw a pixel-wise adjacency between objects moments, which represents distances in terms of pixels.

Unfortunately, this is not an accurate measure since objects might be overlapped and consequently the distance in pixels does not have any significance. For this purpose, we introduce that a paired labeled point between camera and Lidar is the farthest object in the Lidar scan and the farthest one being detected in the background of the image.     

\section{3D Objects from Lidar Point Cloud Data}\label{3DObjectsDetectionFromPointCloudLidarData}
\begin{figure}[H]
\begin{center} 
\includegraphics[width=9cm,height=6cm]{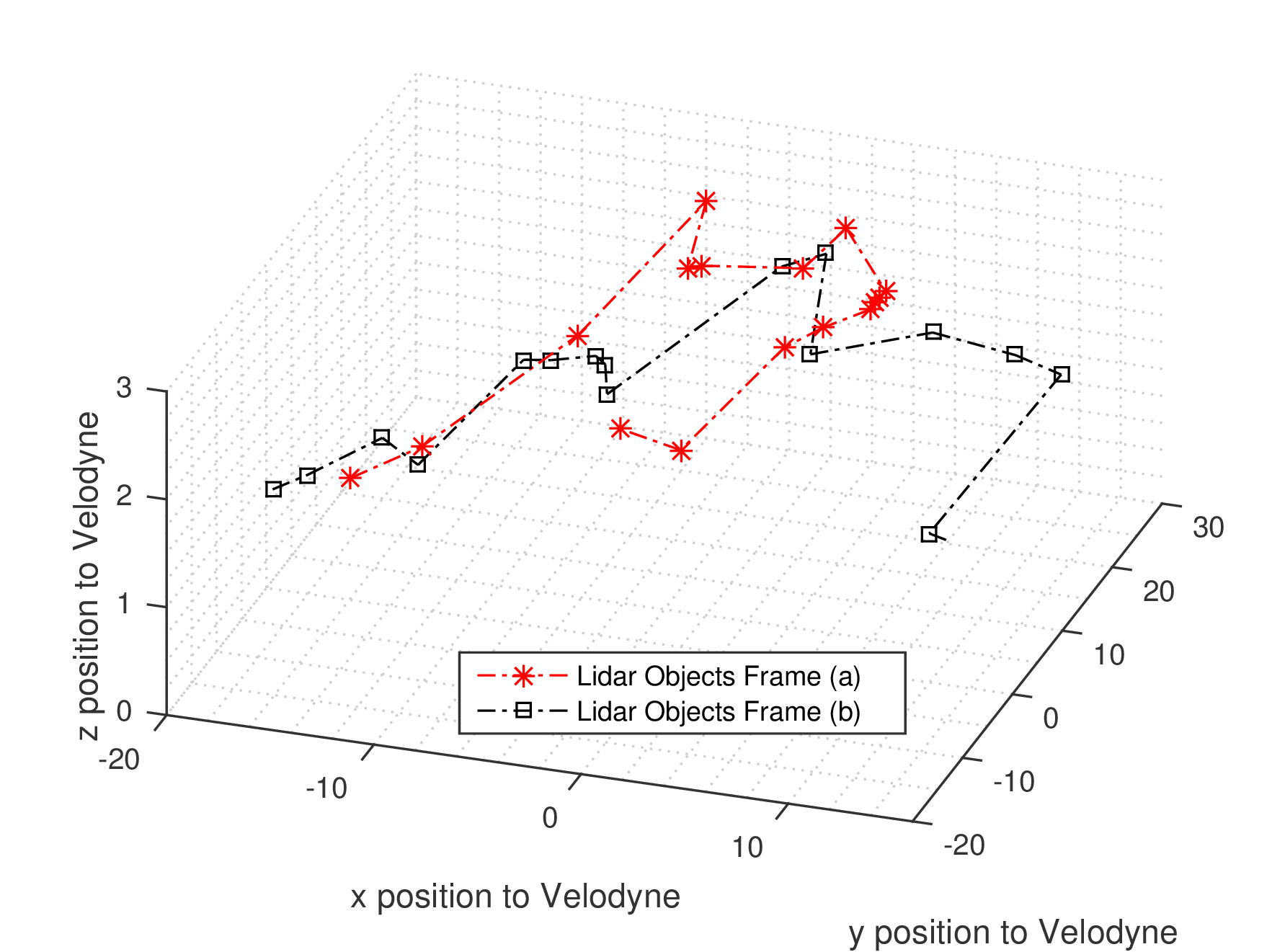}
\end{center}
\caption{Adjacency of Detected Objects from Lidar}
\label{AdjacencyOfDetectedObjectsFromLidar}
\end{figure}

For simplification purposes, we are not considering every object from the Lidar since a tremendous number of unknown objects is detected as a set of neighbored point cloud. In Fig. \ref{AdjacencyOfDetectedObjectsFromLidar}, the dots represents the objects detected from Lidar with \textit{(x,y,z)} triplet that represents relative position to the Velodyne Lidar Scanner. Most of the recognized object classes from Point Cloud data from 3D CNN or labeled data are unknown or do not represent major importance in the alignment. We note that the manifold representing the adjacent objects from Lidar contains larger number of objects comparing to the one from the Camera as in Fig. \ref{PixelWiseAdjacencyOfRecognizedObjects}. For example, the vehicle next to the paired point is not detected in Fig. \ref{FrameB_object_detection}, but is detected in Fig.\ref{AdjacencyOfDetectedObjectsFromLidar} in addition to other objects that are behind of the camera and are captured by Lidar Scans. 
\section{Semi-supervised Alignment of Manifolds: Camera to Lidar and Camera to BSMs} \label{SemiSupervisedAlignmentOfManifolds}
Our problem formulation of manifolds alignment is to be applied to find correspondences between source data containing recognized objects from 2D camera and objects from 3D Lidar point cloud and from 3D V2V exchanged messages. The manifold alignment performs the mapping between the dataset by first successfully learning the low-dimensional embeddings by creating a weighted graph of the objects in the data by finding their correlation  while preserving their neighborhood correlation to preserve the local structure of the data.   
Let $\mathcal{X}$, $\mathcal{Y}$ and $\mathcal{Z}$ three separate data sets consisting respectively of \textit{x,y} and \textit{z} recognized points from camera, Lidar and V2V BSMs. 
We consider consider creating three different Laplacian graphs for each data sets $\mathcal{X}$, $\mathcal{Y}$ and $\mathcal{Z}$. 
The neighborhood weights of a point $t^{(i)}$ as a node in every Laplacian graph for each data set by solving the following optimization problem: \\
\begin{equation} \label{SystemStates}
     \arg\min_{W_{ij}} = \left\{ \left|\textbf{t}^{(i)}- \sum_{j\in\mathcal{N}(i)} W_{ij} \textbf{t}^{(\mathcal{N}(i,j))}\right|^2 \right\} \\
     \textbf{s.t}  \sum_{j\in\mathcal{N}(i)} W_{ij}=1
\end{equation}

The distance matrix $D_{t}$ of point $t^{(i)}$ in one of those data sets to its nearest neighbors points in the same data set characterizes the value of the weight and can be represented as: \\
\begin{equation}
D_{t}=\left[\begin{matrix} t^{(i)}-t^(\textit{N(i,1}) \\
                       t^{(i)}-t^(\textit{N(i,2})     \\ 
                       .\\
                       .\\
                       .\\
                       t^{(i)}- t^(\textit{N(i,N}) \\
      \end{matrix}\right]
\end{equation}
The neighborhood importance between a point $t^{(i)}$ to another point t\^{($\mathcal{N}$(i)}) is proportional with the value on the edge between them, in the new low space representation.
\begin{equation}
W_{ij} = \ddfrac{\sum_{k=1}^{N} \left\{(\textbf{D}_i \textbf{D}^{T}_i)^{-1}\right\}_{jk}}{\sum_{m=1}^{N}\sum_{n=1}^{N}\left\{(\textbf{D}_i \textbf{D}^{T}_i)^{-1}\right\}_{mn}}
\end{equation}
We note that ${(D_{i}D_{i}^{T})^{-1}}$ is the element of the \textit{u}th row and the \textit{v}th column in the inverse of matrix $(D_{i}D_{i}^{T})$.

The problem of the alignment between the camera source data $\mathcal{X}$ to the destination Lidar data $\mathcal{Y}$ and separately to the $\mathcal{Z}$ destination DSRC data set can be expressed as:
\begin{equation}
\begin{split}
\arg\min_{\text{f,g}}\left\{\lambda^x \sum_{i,j}[f_i - f_j]^2 W^x_{ij} + \lambda^y \sum_{i,j}[g_i - g_j]^2 W^y_{ij} \right. \\ \left.  + \mu \sum_{i \in \mathcal{P}}|f_i - g_i|^2 \right\} ~~~~~~~~~~~~~~~~~~~~~   
\end{split}
\label{argminManifoldAlignment1}
\end{equation}

Where $f=[f_{1}, .., f_{x}]^{(T)}$ are vectors in $R^{(x)}$ of the \textit{x} camera points, while $g=[g_{1}, .., g_{y}]^{(T)}$ are vectors in $R^{(y)}$ of the \textit{y} Lidar points and $g=[g_{1}, .., g_{z}]^{(T)}$ for \textit{z} V2V points. For the different parts, we use the following wighting factors $\lambda^{x}$, $\lambda^{y}$ and $\mu$.
We consider that $\mathcal{P}$ is the the set that contains the indices of paired points between camera and Lidar for one alignment and the camera and the V2V BSMs for the other alignment, as shown in Fig \ref{PixelWiseAdjacencyOfRecognizedObjects}. We minimize the first term in order to have the larger $W_{ij}^{x}$ gets the smaller $f_{i}-f_{j}$ is, which guarantees the preservation of the neighbourhood relations of $\mathcal{X}$ camera data set within the elements of $f$. Same thing applies to $g$ for both $\mathcal{Y}$ Lidar dataset and $\mathcal{Z}$ V2V BSMs dataset, while minimizing the second term. The final term in Eq. (\ref{argminManifoldAlignment1}) has an effect to penalize discrepancies between the paired points in the $f$ vector from $\mathcal{X}$ and $g$ vector from both $\mathcal{Y}$ and $\mathcal{Z}$. 
Eq. (\ref{argminManifoldAlignment1}) can be reformulated as: 

\begin{equation}
\arg\min_{\text{f,g}}\left\{\lambda^x\textbf{f}^TL^x\textbf{f} + \lambda^y\textbf{g}^TL^y\textbf{g} + \mu(\textbf{f} - \textbf{g})^T(\textbf{f} - \textbf{g})        \right\}
\label{argminManifoldAlignment2}
\end{equation}
where $L^x = [L^x_{ij}]~\forall~i,~j \in \mathcal{X}$, such that:

\begin{equation}
L_{ij}^x = 
\begin{cases}
\sum_{j} W_{ij}^x, & i = j \\
-W^x_{ij} & j \in \mathcal{N}_i\\
0 &\text{Otherwise}
\end{cases}
\label{Lijx}
\end{equation}

while $L^{y}= [L^y_{ij}]$ used for camera-Lidar $\forall i,j \in \mathcal{Y} $  and $L^{z}= [L^z_{ij}]$ used for camera-V2V $\forall i,j \in \mathcal{Z}$.
A hard constraint is imposed so that $f_{i}=g_{i}$ (i.e. equal eigen vectors for same index of paired points) $\forall i \in \mathcal{P}$ (i.e. as $\mu \to \infty$).
The problem in (\ref{argminManifoldAlignment2}) is transformed into an eigenvalue problem as follows:\\
\begin{equation}
\arg\min_{\text{h}}\left\{\frac{\textbf{h}^TL^z\textbf{h}}{\textbf{h}^T\textbf{h}}   \right\}
\label{argminHTLZ}
\end{equation}

\begin{equation}
\text{s.t.} h^{T}1=0
\end{equation}

\begin{equation}
\textbf{h} = \left[ \begin{matrix} \textbf{f}_{\mathcal{P}} =  \textbf{g}_{\mathcal{P}} \\ \textbf{f}_{\mathcal{Q}^x} \\  \textbf{g}_{\mathcal{Q}^y}
\end{matrix} \right]
\end{equation}
with $\mathcal{Q}^{x}= \mathcal{X}/\mathcal{P}$ and $\mathcal{Q}^{y}= \mathcal{Y}/\mathcal{P}$ and with taking differences between the two seperate alignments that are applied either in camera-Lidar or camera-V2V BSMs alignment. 

\begin{equation}
L^z=
  \begin{bmatrix}
    \lambda^xL^x_{\mathcal{P}\mathcal{P}} + \lambda^yL^{y}_{\mathcal{P}\mathcal{P}} & \lambda^xL^x_{\mathcal{P}\mathcal{Q}^x} & \lambda^yL^y_{\mathcal{P}\mathcal{Q}^y} \\
    \lambda^xL^x_{\mathcal{Q}^x\mathcal{P}} & \lambda^xL^x_{\mathcal{Q}^x\mathcal{Q}^x} & \textbf{0} \\
    \lambda^yL^y_{\mathcal{Q}^y\mathcal{P}} & \textbf{0} & \lambda^yL^y_{\mathcal{Q}^y\mathcal{Q}^y}
  \end{bmatrix}
\end{equation}
while having $L^{x}_{\mathcal{I}\mathcal{J}}$ $(L^{y}_{\mathcal{I}\mathcal{J}} ~ or~ L^{z}_{\mathcal{I}\mathcal{J}}) $ is the sub-matrix of the the matrix $L^{x} (L^{y}~or~L^{z})$ depending on both of types of the alignment.
The solution of the alignment problem is the eigenvector \textbf{h} that corresponds to the smallest non-zero eigenvalue of $L^{z}$. In addition, \textbf{h} is structured in a way that it begins with the $\mathcal{P}$ paired points of f and g, then it is followed by the remaining data points of \textbf{f} and ends with rest of points of \textbf{g}. 
Depending on the selected \textit{l} dimensional embedding chosen for both of the alignment, we will end up with different \textit{l} eigenvectors $[h^{(\textit{1})},..,h^{(\textit{l})}]$ after each joint graph Laplacian for the two alignment process. 
The structure of the embedding of the points of camera-Lidar points and camera-V2V BSMs points are contained in two different matrices that characterize the expected neighborhoods between the points of the datasets and is given by :\\
\begin{equation}
\mathcal{E}=\left[\begin{matrix} f_{\mathcal{P}}^{(i)} &.&.&. ~~ f_{\mathcal{P}}^{(\textit{l})} \\
                       f_{\mathcal{Q}^{(x)}}^{(\textit{1})} &.&.&. ~~ f_{\mathcal{Q}^{(x)}}^{(\textit{1})}\\
                       f_{\mathcal{Q}^{(y)}}^{(\textit{1})} &.&.&. ~~ f_{\mathcal{Q}^{(y)}}^{(\textit{l})}\\
      \end{matrix}\right]
\end{equation}
\section{Numerical Analysis of Mapping Accuracy and Errors} \label{NumericalAnalysisOfMappingAccuracyAndErrors}
\begin{figure}[H]
\begin{center} 
\includegraphics[width=9cm,height=6cm]{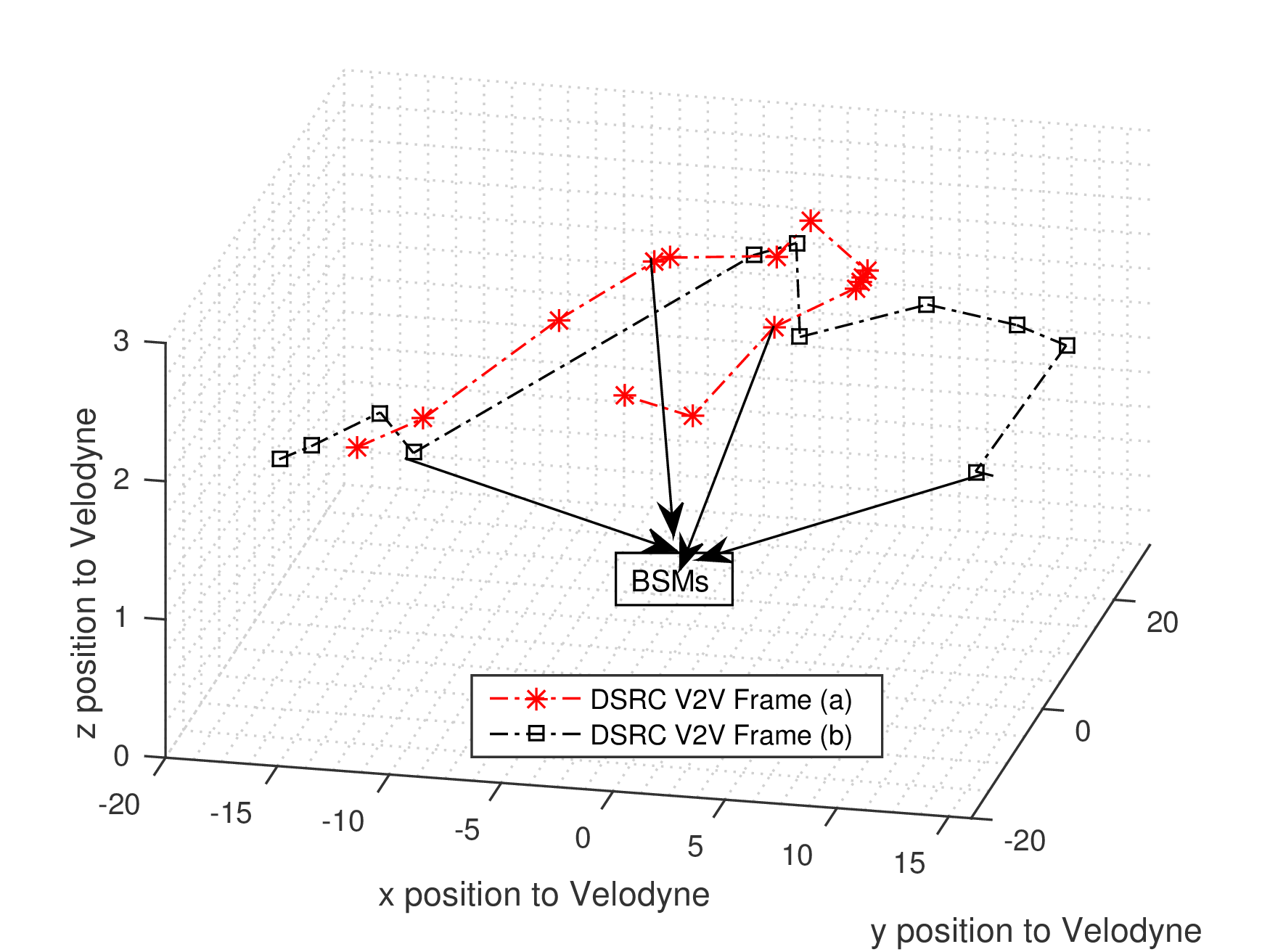}
\end{center}
\caption{Adjacency of Recognized Objects from DSRC}
\label{AdjacencyOfRecognizedObjectsFromDSRC}
\end{figure}

\begin{figure}[H]
\begin{center} 
\includegraphics[width=9cm,height=7cm]{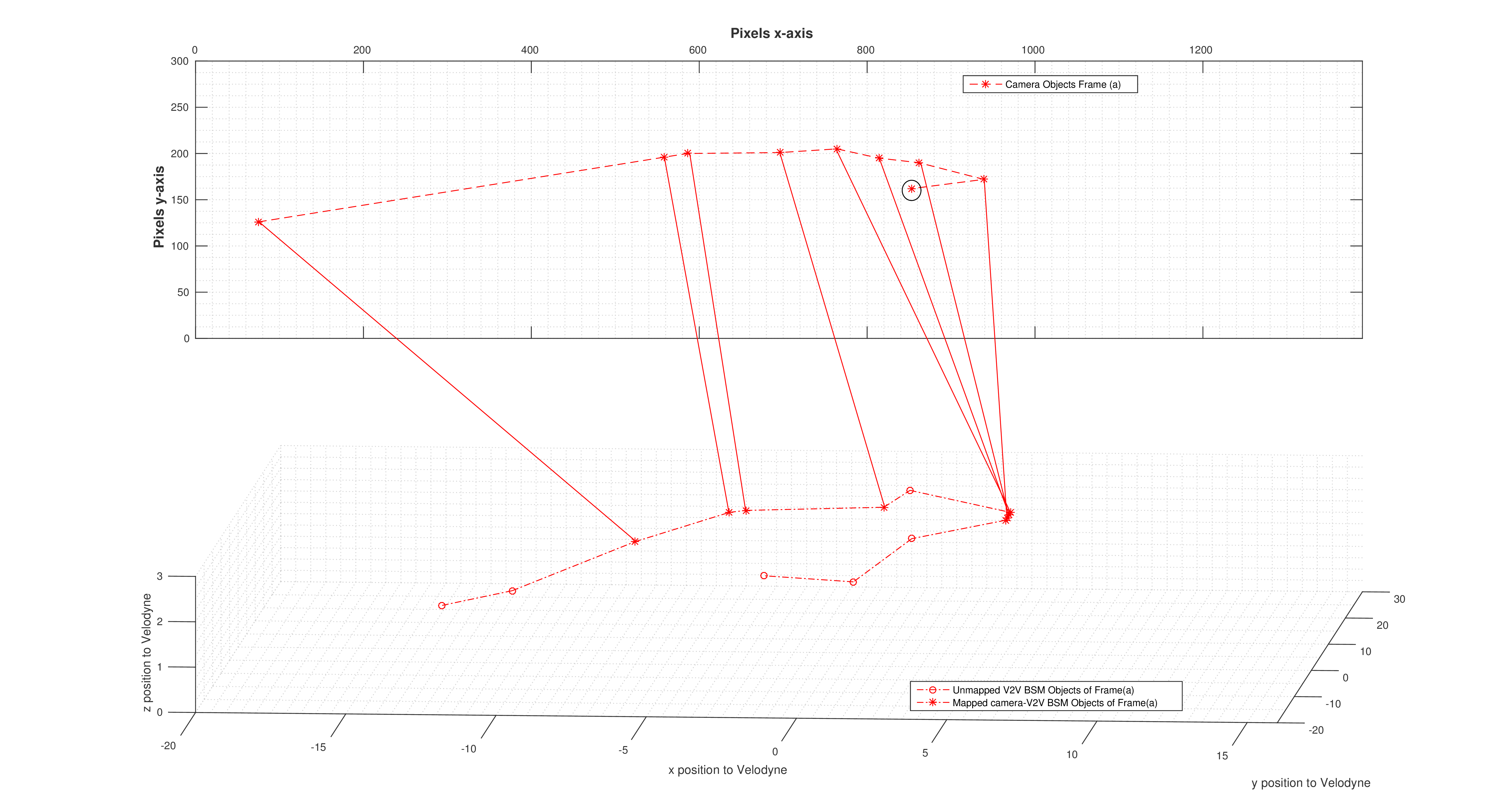} 
\end{center}
\caption{Camera-V2V Object Alignment frame (a)}
\label{Camera2DSRCObjectAlignment_frame_a}
\end{figure}

\begin{figure}[H]
\begin{center} 
\includegraphics[width=9cm,height=6cm]{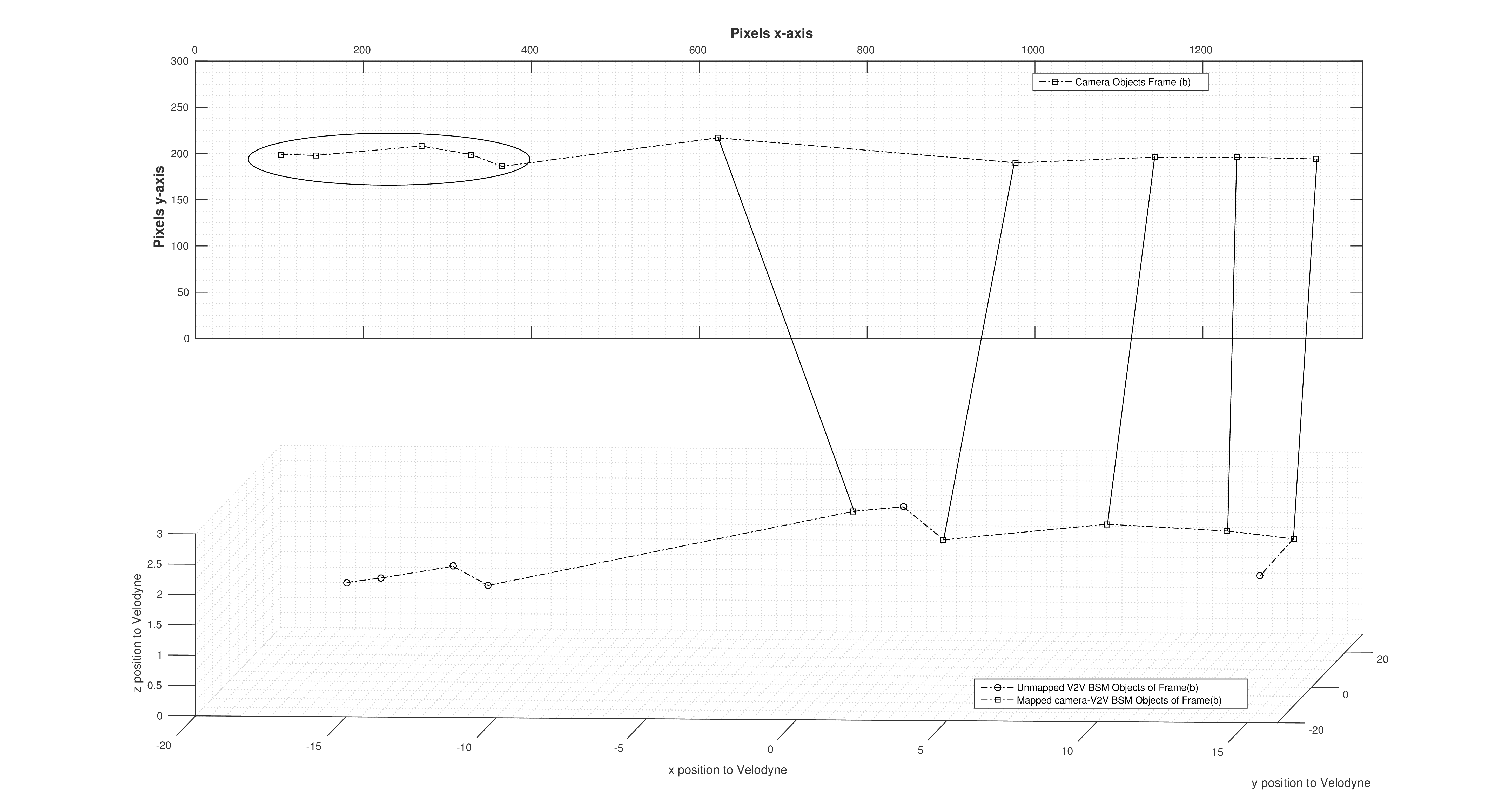}
\end{center}
\caption{Camera-V2V Object Alignment frame (b)}
\label{Camera2DSRCObjectAlignment_frame_b}
\end{figure}

\begin{figure}[H]
\begin{center} 
\includegraphics[width=9cm,height=6cm]{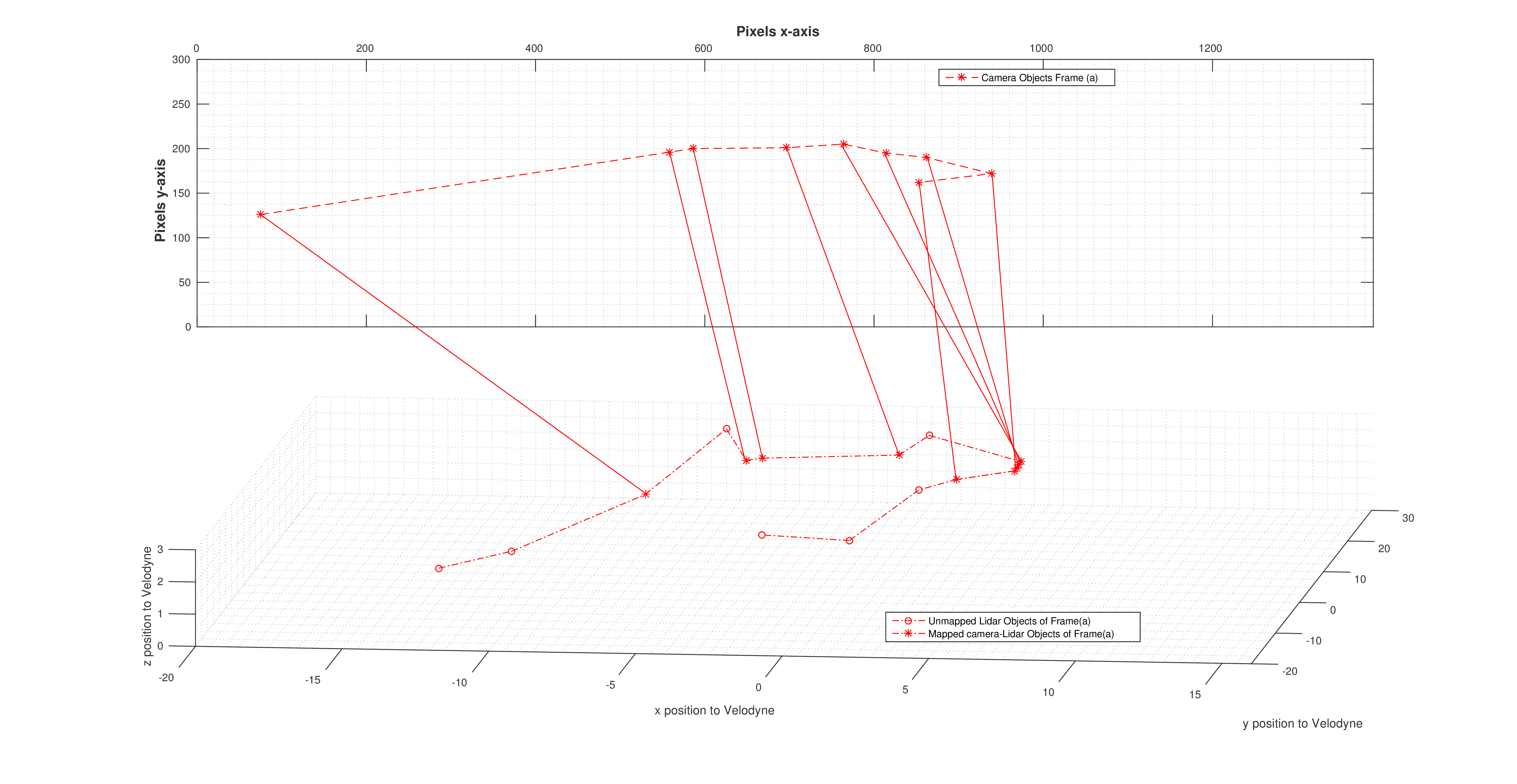}
\end{center}
\caption{Camera-Lidar Object Alignment frame (a)}
\label{Camera2LidarObjectAlignment_frame_a}
\end{figure}

\begin{figure}[H]
\begin{center} 
\includegraphics[width=9cm,height=6cm]{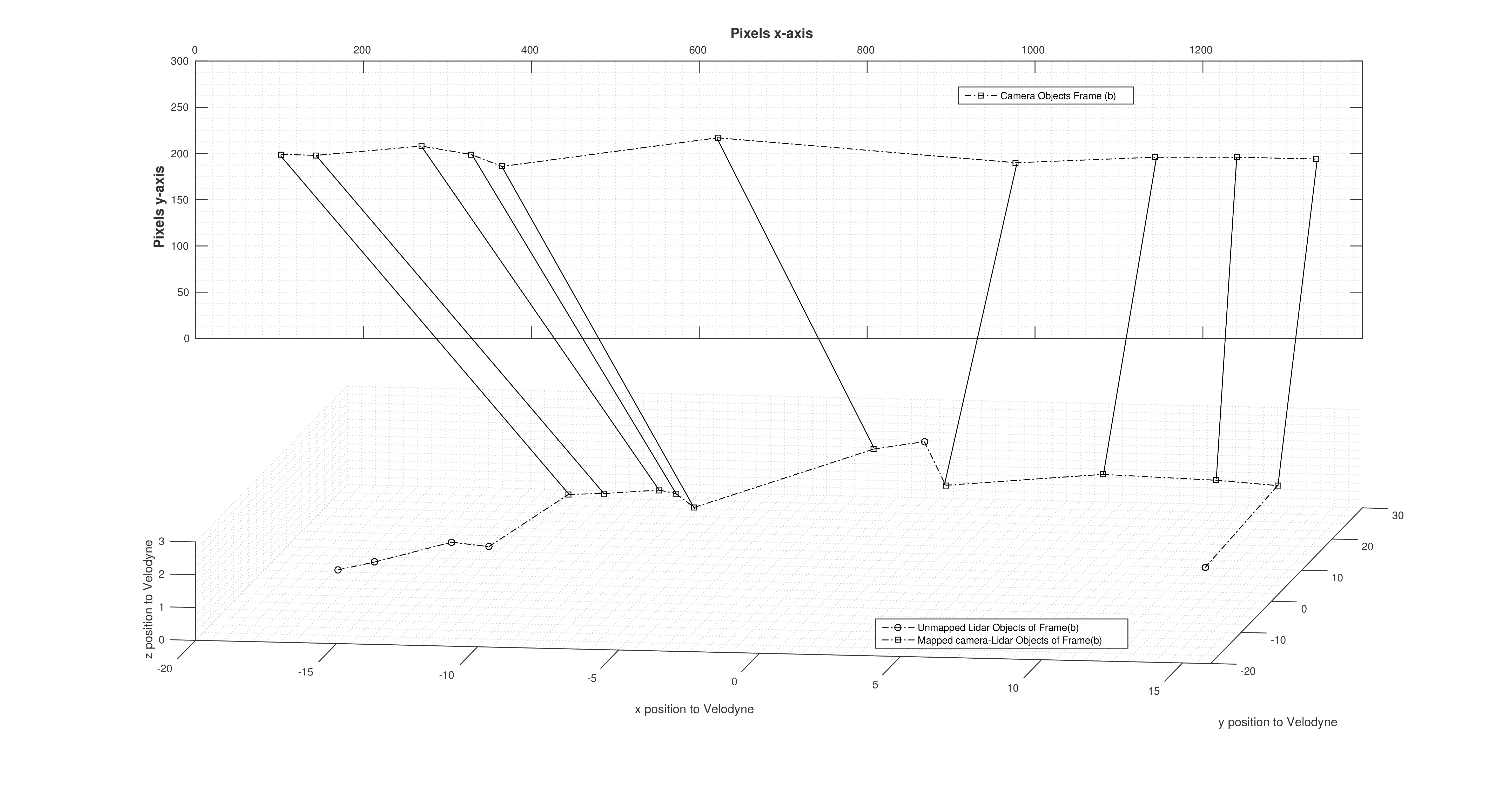}
\end{center}
\caption{Camera-Lidar Object Alignment frame (b)}
\label{Camera2LidarObjectAlignment_frame_b}
\end{figure}

The lack of timestamped V2V BSMs data that corresponds to camera frames and Lidar scans, have encouraged us to dynamically generate the V2V related messages of the recognized objects from the Lidar as shown in Fig. \ref{AdjacencyOfRecognizedObjectsFromDSRC}. We consider that the only class of objects that generate BSM are cars. We note that the generated BSMs are simple and do not include the following fields, such as messages count, temporary ID, brake system status and acceleration set 4 way. For position accuracy, we consider using the position \textit{(x,y,z)} triplet of the object to the Velodyne Lidar Scanner as a replacement to the real-world positioning from \textit{(Latitude, Longitude, Elevation)}. We notice that the shape and the number of points in the manifold representing the V2V BSMs objects in Fig. \ref{AdjacencyOfRecognizedObjectsFromDSRC} is different than the one formed by Lidar objects and presented in Fig. \ref{AdjacencyOfDetectedObjectsFromLidar}. This is related to the number of objects of class person that do not generate V2V BSMs but were reported by Lidar.
We are considering a semi-supervised technique to achieve the alignment of different set of objects that are characterized by the same underlying manifold in 2D camera with 3D Lidar and 2D camera with 3D V2V BSMs. We are considering the frames that have at least one easily paired object and that corresponds to the largest distance from the Lidar and the one in the background of the image.  \begin{figure}[H]
\begin{tikzpicture}
\pgfplotsset{width=9cm,height=6cm}
\begin{axis}[
	x tick label style={
		/pgf/number format/1000 sep=},
	ylabel=Number of Objects,
	xlabel= \\ Class(Frame),
	symbolic x coords={CAR(a),PERSON(a),CAR(b),PERSON(b)},
    xtick=data,
	enlargelimits=0.15,
	legend style={at={(0.5,1.1)},
		anchor=north,legend columns=-1},
	ybar,
	bar width=2pt,
	grid=major,
	symbolic y coords={0,1,2,3,4,5,6,7,8,9,10,11,12,13,14,15},
	ytick={0,1,...,15},
]
\addplot[ybar,color=blue,fill=blue] coordinates {
        (CAR(a),14)
        (PERSON(a),2)
        (CAR(b),11)
        (PERSON(b),5)
    };
\addplot[ybar,color=red] coordinates {
        (CAR(a),8)
        (PERSON(a),1)
        (CAR(b),5)
        (PERSON(b),5)
        
    };
\addplot[ybar,color=green,postaction={pattern=north east lines}] coordinates {
        (CAR(a),14)
        (PERSON(a),0)
        (CAR(b),11)
        (PERSON(b),0)
    };
\legend{Lidar,Camera,DSRC}
\end{axis}
\end{tikzpicture}
\caption{Number of Objects per Class and Frame }%
\label{VCCVMsUtilizationForGreedyAndMDPSchemes}
\end{figure}
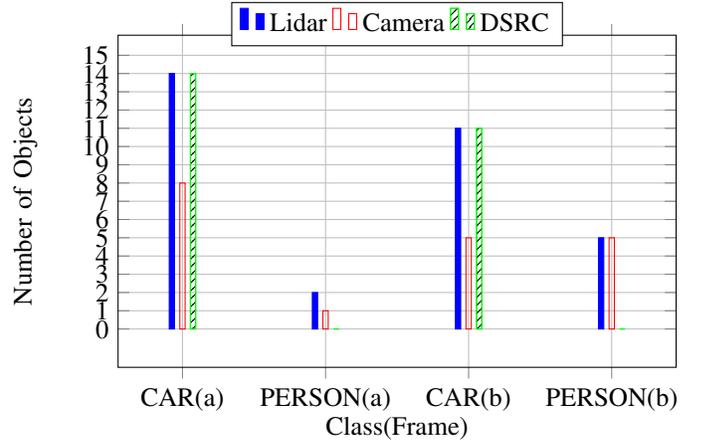
The correspondences of every pair of points between both the camera to Lidar and the camera to V2V BSMs are given by the lowest value in the final structure of the embedding $\mathcal{E}$ matrix.
We define the overall  percentage error in mapping between all all the points from $\mathcal{X}$ to $\mathcal{Y}$ and to $\mathcal{Z}$ by the number of erroned correspondences over the number of points of $\mathcal{X}$ camera data set, as given by the equation:
\begin{equation}
\begin{split}
    e = \sum_{i=1}^{x} \frac{c_{ij}}{x} ~~~~~~~~~~~~\\
    \text{where}~ c_{ij}=
    \begin{cases}
        0   & x_{i}=y_{j} ~||~ x_{i}=z_{j}  \\
        1   & Otherwise,
    \end{cases}
\end{split}    
\end{equation}
$c_{ij}$ characterizes if the expected point $y_{j}$ from Lidar data set $\mathcal{Y}$  or $z_{j}$ from V2V BSMs data set $\mathcal{Z}$ corresponds to the point $x_{i}$ from the camera data set  $\mathcal{X}$.    
Fig. \ref{Camera2DSRCObjectAlignment_frame_a} and Fig. \ref{Camera2DSRCObjectAlignment_frame_b} presents the results of the alignment between camera objects to the corresponding BSMs objects respectively for frame\textit{(a)} and frame\textit{(b)} . 
The circled points in Fig. \ref{Camera2DSRCObjectAlignment_frame_a} and Fig. \ref{Camera2DSRCObjectAlignment_frame_b} represents the objects from camera that were not aligned with their correspondances in V2V BSMs data set. Enriched points in the V2V BSMs objects that do not have correspondences are enriching the 3D reconstruction of the scene. These objects are either out-of-sight of the camera or that were not recognized. 
Fig. \ref{Camera2LidarObjectAlignment_frame_a} and Fig. \ref{Camera2LidarObjectAlignment_frame_b} represents the alignment between the objects of the camera objects respectively from frame\textit{(a)} and frame\textit{(b)} with corresponding objects of the Lidar point cloud. 
Specifically, the points in circles were not mapped to any of those from the camera since they have low correlation with the neighboring points and are additive points from Lidar point cloud for both of the frames corresponding to both undetected or out-of-sight objects that were not detected by the frames.  We notice that the mapping of the objects recognized from the camera to the one from Lidar is more accurate and present more matching than to the V2V objects. 
The enriched objects per technology after the mapping for both of the scene is given by the Table \ref{DataEnrichedObjects}. It represents the objects of the autonomous vehicle scene that are specific to the technology and that were not aligned between the data sets. 
\begin{table}[ht!]\small
\centering
\caption{\label{DataEnrichedObjects} Unmapped Objects from the 3 different modalities of sensors}
\begin{tabular}{p{1.5cm} r lr lr lr }
\toprule
Data set & Classe &   \multicolumn{1}{c}{Frame(a)} &  \multicolumn{1}{c}{Frame(b)} \\
\midrule
\multirow{2}{1.5cm}{Camera} 
&Car ~~~&    ~~~~~~~0 & 0~~~~~~~ &\\
&Person &    ~~~~~~~1 & 5~~~~~~~ &\\
\midrule
\multirow{2}{1.5cm}{Lidar}
&Car ~~~&    ~~~~~~~6 & 6~~~~~~~ &\\
&Person &    ~~~~~~~1 & 0~~~~~~~ &\\
\midrule
\multirow{2}{1.5cm}{V2V BSMs}
&Car ~~~&    ~~~~~~~6 & 6~~~~~~~ &\\
&Person &    ~~~~~~~0 & 0~~~~~~~ &\\
\bottomrule
\end{tabular}
\end{table}
\section{Conclusion}\label{Conclusion}
We developed a framework for corresponding between objects recognized from camera data set to Lidar and to DSRC data sets that are characterized with the same underlying manifold. The mapping allows us to be more informed about one object that was paired between one or two data sets in terms of texture details from picture, V2V communication details as well as the 3D shape and accurate distance to surroundings. 3D point cloud, 2D images and V2V 3D information are supplementary to each other to accomplish robust perception of roads. Full scene reconstruction of objects from the three input modes is represented with focus on the increased added points from each technology. In the future work we plan to study the effect of confusion in each input mode in the 3D scene reconstruction without pre-identified paired points. 
\ifCLASSOPTIONcaptionsoff
  \newpage
\fi
\bibliographystyle{IEEEtran}
\bibliography{IEEEabrv,main.bbl}



\end{document}